\documentclass[a4paper]{report}
\usepackage[utf8]{inputenc}
\usepackage[T1]{fontenc}
\usepackage{RJournal}
\usepackage{amsmath,amssymb,array}
\usepackage{booktabs}
\usepackage{algorithm}
\usepackage{algpseudocode}
\usepackage{amsmath}
\usepackage{xcolor}
\usepackage{makecell}
\usepackage{listings}
\usepackage{tabularx}

\begin{document}

\sectionhead{Contributed research article}
\volume{XX}
\volnumber{YY}
\year{20ZZ}
\month{AAAA}

\newtheorem{defn}{Definition}[section]

\begin{article}

\newcommand{\liveTitle}{\href{https://github.com/MI2DataLab/live}{live}}
\newcommand{\live}{\href{https://github.com/MI2DataLab/live}{live }}
\definecolor{orange}{rgb}{1,0.5,0}
\newcommand{\todo}{\textcolor{red}}
\newcommand{\M}{\mathcal{M}}

  \title{Explanations of model predictions with \texttt{live} and \texttt{breakDown} packages
}
\author{by Mateusz Staniak and Przemysław Biecek}

\maketitle

\abstract{
Complex models are commonly used in predictive modeling. In this paper we present R packages that can be used for explaining predictions from complex black box models and attributing parts of these predictions to input features.
We introduce two new approaches and corresponding packages for such attribution, namely \pkg{live} and \pkg{breakDown}. We also compare their results with existing implementations of state-of-the-art solutions, namely \pkg{lime} that implements Locally Interpretable Model-agnostic Explanations and \pkg{ShapleyR} that implements Shapley values.}

\section{Introduction} 

Predictive modeling is a very exciting field with a wide variety of applications. 
Lots of algorithms have been developed in this area. As proven in many Kaggle competitions \citep{kaggle}, winning solutions are often obtained with the use of elastic tools like random forest, gradient boosting or neural networks. 

These algorithms have many strengths but they also share a major weakness, which is deficiency in interpretability of a model structure. A single random forest, an \pkg{xgboost} model or a neural network may be parameterized with thousands of parameters which make these models hard to understand. Lack of interpretability results in the lack of trust in model predictions. Lack of trust is a major obstacle when one thinks about applications in regulated areas such as personalized medicine or similar fields.
An interesting example of a situation in which trust issues are fully justified is presented in \cite{lime}. 
Authors compare two classifiers that were trained to recognize whether a text describes Christianity or Atheism.
After explanations were provided, it turned out that the model with superior performance in the test set often based its prediction on irrelevant words, for example prepositions.
To overcome this problem, the interpretability of complex machine learning models has been a subject of much research, devoted partially also to model visualization. Find some examples in  \cite{Strumbelj:2011:GMV:1997005.1997009, nnet_vis, Zeiler_Fergus_2014}.

The general approach to interpretability is to identify important variables (features) in the model and then learn the expected model response for a single variable. A description of a general framework of permutation-based variable importance rankings may be found in \cite{Altmann_Tolosi_Sander_Lengauer_2010}. 
An interesting and widely adopted tool for estimation of marginal model response are Partial Dependency Plots (see \cite{friedman2001}), that presents the marginal relation between the variable of interest and a single variable from the model. 
An effective and very elastic implementation of this method is available in the \pkg{pdp} package for \texttt{R} (see \citet{pdp}). 
This method has many extensions such as for example Individual Conditional Expectations (\cite{goldstein_peeking_2015}). 
The ICE method allows for tracing predictions for individual variables and it is very useful for identification of interactions. 
On the other hand, ALE plots (\cite{aleplot_art}) were proposed as a superior tool for handling strongly correlated predictors by describing conditional distribution of predicted values. 
This method can be used to assess both main effects and interactions between predictors.
All these methods are focused on the effect of a single variable or small set of variables within the black box model.

A different approach is presented in the article \cite{lime}. 
The authors propose LIME (Locally Interpretable Model agnostic Explanations) as a method for explaining black box predictions by fitting an interpretable model locally around a prediction of interest. 
This methodology was illustrated with examples from image and text classification areas.
Later, it was extended by \cite{magix} to MAGIX methodology (Model Agnostic Globally Interpretable Explanations) and modified by the authors of the original article to aLIME (anchor-LIME) in \cite{alime}.

So far, two implementations of the method have been found.
Python library was developed by the authors of the original article and it is available on GitHub at \url{https://github.com/marcotcr/lime}. 
It works for any text or image classifier as well as for tabular data.
Regression models can be explained using simple linear regression.
The R package \pkg{lime} which is a port to the original Python package is maintained by Thomas Lin Pedersen.
This package works with tabular and text data and handles all models supported by either \pkg{caret} or \pkg{mlr} package and it can be easily extended to work with other models.
Neither of the packages implements \textit{sp-LIME} algorithm that was proposed in the original article to choose representative observations that would explain the behavior of the model globally.

In this article we give a short overview of methods for explaining predictions made by complex models. We also introduce two new methods implemented in R packages \pkg{live} and \pkg{breakDown}.

\section{Related work}

In this section we present two most recognized methods for explanations of a single prediction from a complex black box model.

\subsubsection{Locally Interpretable Model-agnostic Explanations (LIME)}

In 2016 \cite{lime} proposed \textbf{LIME} method for explaining prediction for a single observation which takes a significantly different approach compared to the methods described above.
The algorithm is designed to explain predictions of any classifier and it works primarily for image and text data.
First, original observation of interest is transformed into simplified input space of binary vectors (for example presence or absence of words).
Then a dataset of similar observations is created by sampling features that are present in the representation of the explained instance.
Closeness of these observations to the original observations is measured via specified similarity kernel.
This distance is taken into account while the explanation model is fitted to the new dataset.
The interpretable model can be penalized to assure that it does not become too complex itself.

\subsection{Shapley values (SHAP)}

In 2017 \cite{shapley} introduced a general framework for explaining machine learning models that encompasses LIME among other methods.
The method is associated with some specific visualization techniques that present how predictors 
contribute to the predicted values.
In this framework, observations are transformed into the space of simplified inputs.
Explanation models are restricted to the so-called \textbf{additive feature attributions methods}, what means that values predicted by the explanation model are linear combinations of these binary input vectors.
Formally, if $z = (z_{1}, \ldots, z_{p})$ is a vector in simplified inputs space and $g$ is the explanation model, then
$$
g(z) = \phi_{0} + \sum_{j = 1}^{M}\phi_{j}z_{j},
$$
where $\phi_{j}, j = 0, \ldots, M$ are weights.
These weights measure how each feature contributes to the prediction. 
Authors prove that in this class of explanation models \textbf{Shapley values} provide unique solutions to the problem of finding optimal weights $\phi_{j}$ that assure that the model has desirable properties of local accuracy and consistency.
For formal treatment and examples, please refer to the original article \cite{shapley}.
SHAP values cannot currently be computed using any R package available on CRAN, but a development version of the package can be found on \url{https://github.com/redichh/ShapleyR}.

\section{Local Interpretable Visual Explanations (LIVE)} 

The next two sections introduce two alternative approaches to explaining model predictions. Both of them are implemented in R packages, \pkg{live} and \pkg{breakDown}, respectively.

\subsection{Motivation}
 
\live is an alternative implementation of LIME for regression problems, which emphasizes the role of model visualization in understanding of complex models.
In comparison with the original LIME, both the method of local exploration and handling of interpretable inputs are changed.
Dataset for local exploration is simulated by perturbing the explained instance one feature at a time. The process is described in section \ref{algs}.
All generated observations are treated as similar to the observation of interest and thus the identity kernel is used.
Original variables are used as interpretable inputs.
Interpretability of the local explanation comes from a tractable relationship between inputs and the predicted response. 
Variable selection is optional for linear regression when sparsity is required.

One of the main purposes of \live is to provide tools for model visualization, which is why in this package emphasis is put on models that are easy to visualize.
For linear models, waterfall plots can be drawn to present how predictors contribute to the overall model score for a given prediction, while forest plots \citep{forestplot} can be drawn to summarize the structure of local linear approximation.
Examples clarifying both techniques are given in section \ref{case_study}.
Other interpretable models that are equipped with generic \texttt{plot} function can be visualized, too.
In particular, decision trees which can be plotted using \pkg{partykit} package are well suited for this task, as they can help discover interactions.
The package uses the \texttt{mlr} interface to handle machine learning algorithms, hence any classifier or regression method supported by \texttt{mlr} can be used as a black box model and similarly any classifier or regression method can be used as an interpretable model.

\subsection{Methodology}
\label{algs}
\pkg{live} package uses a two-step procedure to explain prediction of a selected black box model in the point $x$. First, an artificial dataset $X'$ is created around point $x$. Second, the white box model is fitted to the model predictions in points $X'$.

The first step is described by the Algorithm 1. 
When the number of predictors is smaller than or equal to the desired size of the simulated dataset for local exploration, it is created in accordance with the following procedure.
\begin{algorithm}
\caption{Simulating $X'$ - surroundings around the selected $x$.}
\begin{algorithmic}[1]
\State $p \gets \text{number of predictors}$
\State $n \gets \text{number of observations to generate}$
\State $\text{Duplicate the given observation } n \text{times}$
\If{$p > n$}
\State{Randomly pick n predictors}
\For{i in $\{1, \ldots, n\}$}
\State{Draw number $k \in \{1, \ldots, n\}$ uniformly. Replace $k$-th variable in I with a random draw from the empirical distribution of this variable [in i-th duplicate of the original observation]}
\EndFor
\Else
\For{i in $\{1, \ldots, p\}$}
\State{Replace the value of i-th variable in i-th observation with a draw from empirical distribution of this variable}
\State{For the remaining n - p observations, proceed as in case $p > n$.}
\EndFor
\EndIf
\end{algorithmic}
\end{algorithm}
When the number of predictors is bigger than the size of \textit{fake} dataset, a random subset of $n$ predictors is chosen, where $n$ is the number of observations, and the procedure described above is performed on this subset.
In other words, the procedure amounts to iterating over the set of $n$ observations identical to a given instance and changing the value of one variable at each step. 
Current implementation of this algorithm relies on \pkg{data.table} package for performance \citep{data_table}.

\section{Model agnostic greedy explanations of model predictions (breakDown)} 

\subsection{Motivation}

\pkg{live} package approximates the local structure of the black box model around a single point in the feature space. The idea behind the \pkg{breakDown} is different. In case of that package the main goal is to decompose model predictions into parts that can be attributed to particular variables. It is straightforward for linear (and more general: additive) models. Below we present a model agnostic approach that works also for nonlinear models.

Let us use the following notation: $x = (x_1, x_2, ..., x_p) \in X \subset \mathcal R^p$ is a vector in feature space $X$. $f:X \rightarrow R$ is a scoring function for the model under consideration, that may be used for regression of classification problems. $X^{train}$ is a training dataset with $n$ observations.

For a single observation $x^{new}$ the model prediction is equal to $f(x^{new})$. Our goal is to attribute parts of this score to variables (dimensions) in the $X$ space. 

\subsection{The lm-break: version for additive models}

For linear models (and also generalized linear models) the scoring function (e.g. link function) may be expressed as linear combination of feature vectors.

\begin{equation}
f(x^{new}) = (1, x^{new}) (\mu, \beta)^T = \mu + x^{new}_1 \beta_1 + \ldots + x^{new}_p \beta_p.
\label{eq:linearmodel}
\end{equation}

In this case it is easy to attribute the impact of feature $x_i$ to prediction $f(x^{new})$. The most straightforward approach would be to use the $x^{new}_i \beta_i$ as the attribution. However, it is easier to interpret variable attributions if they are invariant to scale-location transformations of $x_i$, such as change of the unit or origin.  This is why for linear models the \textbf{lm-break} variable attributions are defined as  $(x^{new}_i - \bar x_i) \beta_i$. The equation \ref{eq:linearmodel} may be rewritten as follows:

\begin{equation}
f(x^{new}) = (1, x^{new}) (\mu, \beta)^T = baseline + (x^{new}_1 - \bar x_1) \beta_1 + ... + (x^{new}_p - \bar x_p) \beta_p
\end{equation} 

where
$$
baseline = \mu + \bar x_1 \beta_1 + ... + \bar x_p \beta_p.
$$

Components $(x^{new}_i - \bar x_i) \beta_i$ are all expressed in the same units. For \texttt{lm} and \texttt{glm} models these values are calculated and plotted by the generic \texttt{broken()} function from the \pkg{breakDown} package.

\subsection{The ag-break: model agnostic approach}

Interpretation of \textbf{lm-break} attributions is straightforward, but limited only to additive models. In this section we present an extension for non-additive models. This extension uses additive attributions to explain predictions from non-additive models thus some information about the model structure will be lost. Still, for many models such attribution may be useful.
For additive models the \textbf{ag-break} approach gives the same results as \textbf{lm-break} approach. 

The intuition behind \textbf{ag-break} approach is to identify components of $x^{new}$ that cannot be changed without a significant change in the prediction $f(x^{new})$. In order to present this approach in a more formal way, we first need to introduce some definitions.

\begin{defn}[Relaxed model prediction]
Let $f^{IndSet}(x^{new})$ denote an expected model prediction for $x^{new}$ relaxed on the set of indexes $IndSet \subset \{1, \ldots, p\}$.
\begin{equation}
f^{IndSet}(x^{new}) = E[f(x)|x_{IndSet} = x^{new}_{IndSet}].
\end{equation}
Thus $f^{IndSet}(x^{new})$ is an expected value for model response conditioned on variables from set $IndSet$ in such a a way, that $\forall_{i\in IndSet} x_i = x^{new}_i$.
\end{defn}

the intuition behind relaxed prediction is that we are interested in an average model response for observations that are equal to $x^{new}$ for features from $IndSet^C$ set and follow the population distribution for features from $IndSet$ set. Clearly, two extreme cases are
\begin{equation}
f^{\{1, \ldots, p\}}(x^{new}) = f(x^{new}),
\end{equation}
which is the case of no relaxation, and
\begin{equation}
f^{\emptyset}(x^{new}) = E [f(x)].
\end{equation}
which corresponds to full relaxation.
We will say that a variable was relaxed, when we do not fix its value and we let it follow the population distribution.
This will play a crucial part in the algorithm presented in this section.

Since we do not know the joint distribution of $x$, we will use its estimate instead. 
\begin{equation}
\widehat {f^{IndSet}(x^{new})} = \frac 1n \sum_{i = 1}^n f(x^i_{-IndSet},x^{new}_{IndSet}).
\end{equation}
We will omit the dashes to simplify the notation.

\begin{defn}[Distance to relaxed model prediction]
Let us define the distance between model prediction and relaxed model prediction for a set of indexes $IndSet$.
\begin{equation}
d(x^{new}, IndSet) := |f^{IndSet}(x^{new}) - f(x^{new})|.
\end{equation}
\end{defn}
It is the difference between model prediction for observation $x^{new}$ and observation relaxed on features $indSet$. 
The smaller the difference, the less important are variables in the $indSet$ set.

\begin{defn}[Added feature contribution] For j-th feature we define its contribution relative to a set of indexes $IndSet$ (\textit{added contribution}) as
\begin{equation}
\text{contribution}^{IndSet}(j) = f^{IndSet \cup \{j\}}(x^{new}) - f^{IndSet}(x^{new}).
\end{equation}
It is the change in model prediction for $x^{new}$ after relaxation on $j$.
\end{defn}

The model agnostic feature contribution is based on distances to relaxed model predictions. In this approach we look for a series of variables that can be relaxed in such a way so as to move model prediction from $f(x^{new})$ to a fully relaxed prediction $E [f(x)]$ (expected value for all model predictions). The order of features in this series is important. But here we use a greedy strategy in which we add features to the $indSet$ iteratively (one feature per iteration) and minimize locally the distance to relaxed model prediction. 

This approach can be seen as an approximation of Shapley values where feature contribution is linked with the average effect of a feature across all possible relaxations. These approaches are identical for additive models. For non-additive models the additive attribution is just an approximation in both cases, yet the greedy strategy produces explanations that are easier to interpret.
It is worth noting that similar decomposition of predictions and measures of contribution for classifiers have been examined in \cite{4407709}.

The greedy search can start from a null set of indexes (then in each step a single feature is being relaxed) or it can start from a full set of relaxed features (then in each step a single feature is removed from the set).
The above approaches are called \textit{step-up} and \textit{step-down}, respectively. They are presented in algorithms \ref{alg:relaxedalg} and \ref{alg:relaxedalg2}. 

\begin{figure}
\includegraphics[width=\textwidth]{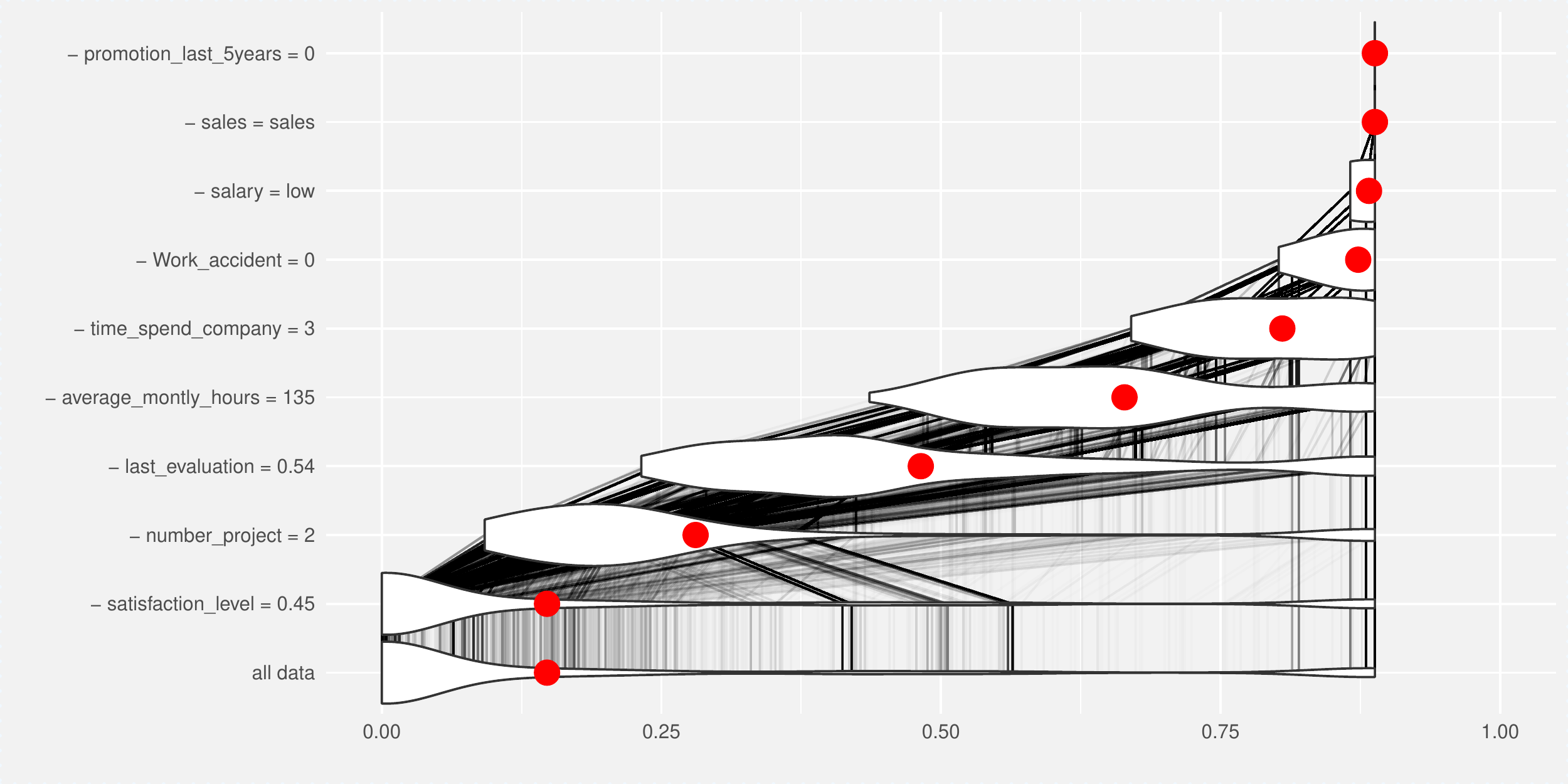}
\caption{\label{fig:breakDownDistr} An illustration of algorithm \ref{alg:relaxedalg}. Each row in this plot corresponds to a distribution of model scores $f(x)|x_{IndSet} = x^{new}_{IndSet}$ for different sets of $IndSet$ indexes. Initially $IndSet = \{1, \ldots, p\}$ and in each step single variable is removed from this set. Labels on the left-hand side of the plots show which variable is removed in a given step. Red dots stand for conditional average - an estimate of relaxed predictions $f^{IndSet}(x^{new})$. Violin plots summarize conditional distributions of scores while gray lines show how model predictions change for particular observations between consecutive relaxations.}
\includegraphics[width=\textwidth]{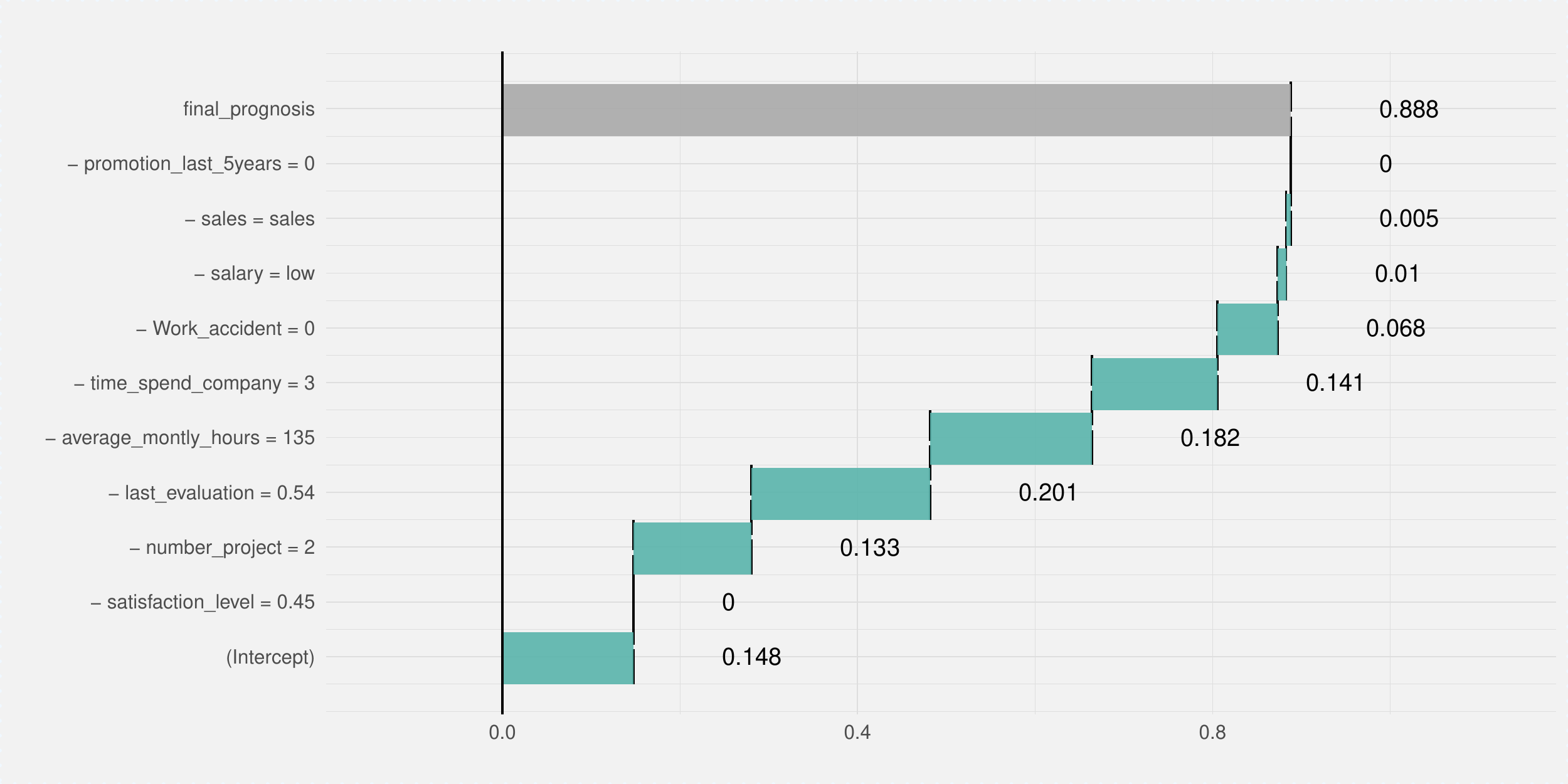}
\caption{\label{fig:breakDownScores} Break Down Plot for decomposition identified in Figure \ref{fig:breakDownDistr}. Beginning and end of each rectangle corresponds to relaxed prediction (red dot in Figure \ref{fig:breakDownDistr}) with and without particular feature.}
\end{figure}

The algorithm \ref{alg:relaxedalg} presents the procedure that generates a sequence of variables with increasing contributions. This sequence corresponds to variables that can be relaxed in such a way so as to minimize the distance to the original prediction. The resulting sequence of $Contributions$ and $Variables$ may be plotted with Break Down Plots, see an example in  Figure \ref{fig:breakDownScores}. Figure \ref{fig:breakDownDistr} summarizes the idea behind algorithm \ref{alg:relaxedalg}. By relaxing consecutive variables one finds a path between single prediction and average model prediction.

One can also consider an opposite strategy - instead of starting from $IndSet = \{1, \ldots, p\}$ one can start with $IndSet = \emptyset$. That strategy is called \textit{step-up} approach and it is presented in Algorithm \ref{alg:relaxedalg2}. 

\clearpage

\begin{algorithm}
\caption{\label{alg:relaxedalg}Model agnostic break down of model predictions. The \textit{step-down} approach.}
\begin{algorithmic}[1]
\State $p \gets \text{number of variables}$
\State $IndSet \gets \{1, \ldots, p\} \text{ set of indexes of all variables}$
\For{i in $\{1, \ldots, p\}$}
\State{Find new variable that can be relaxed with small loss in relaxed distance to $f(x^{new})$}
\For{j in $IndSet$}
\State{Calculate relaxed distance with $j$ removed}
\State{$dist(j) \gets d(x^{new}, IndSet \setminus \{j\})$}
\EndFor
\State{Find and remove $j$ that minimizes loss}
\State{$j_{min} \gets \text{arg}\min_j dist(j)$ }
\State{$Contribution^{IndSet}(i) \gets f^{IndSet}(x^{new}) - f^{IndSet \setminus \{j_{min}\}}(x^{new})$}
\State{$Variables(i) \gets j_{min}$}
\State{$IndSet \gets IndSet \setminus \{j_{min}\}$}
\EndFor
\end{algorithmic}
\end{algorithm}

\begin{algorithm}
\caption{\label{alg:relaxedalg2}Model agnostic break down of model predictions. The \textit{step-up} approach.}
\begin{algorithmic}[1]
\State $p \gets \text{number of variables}$
\State $IndSet \gets \emptyset \text{ empty set}$
\For{i in $\{1, \ldots, p\}$}
\State{Find new variable that can be relaxed with large distance to $f^{\emptyset}(x^{new})$}
\For{j in $\{1, \ldots, p\} \setminus IndSet$}
\State{Calculate relaxed distance with $j$ added}
\State{$dist(j) \gets d(x^{new}, IndSet \cup \{j\})$}
\EndFor
\State{Find and add $j$ that maximize distance}
\State{$j_{max} \gets \text{arg}\max_j dist(j)$ }
\State{$Contribution^{IndSet}(i) \gets f^{IndSet \cup \{j_{max}\}}(x^{new}) - f^{IndSet}(x^{new})$}
\State{$Variables(i) \gets j_{max}$}
\State{$IndSet \gets IndSet \cup \{j_{max}\}$}
\EndFor
\end{algorithmic}
\end{algorithm}

\clearpage

\clearpage

\section{Case study: How good is this red wine?}\label{case_study}


Wine quality data \citep{wine_quality} is a well-known dataset which is commonly used as an example in predictive modeling.
The main objective associated with this dataset is to predict the quality of some variants of Portuguese "Vinho Verde"  on the basis of 11 chemical properties.
A single observation from the dataset can be found in Table \ref{table:example_wine}.
According to the results from the original article, the Support Vector Machine (SVM) model performs better than other models including linear regression, neural networks and others.

In this section we will show how \pkg{live} package can be used to fit linear regression model locally and generate a visual explanation for the black box model as well as how \pkg{breakDown} package can be used to attribute parts of the final prediction to particular features.

\begin{table}[ht]
\centering
\begin{tabularx}{\textwidth}{rrrrrrrrrrrr}
  \hline
\makecell{fixed \\ acidity} & \makecell{volatile \\ acidity} & \makecell{citric \\ acid} & \makecell{res. \\ sugar} & $\text{Cl}^{-}$ & \makecell{free \\ $\text{SO}_{2}$} & \makecell{total \\ $\text{SO}_{2}$} & D & pH & $\text{SO}_{4}^{2-}$ & alcohol \\ 
  \hline
7.40 & 0.70 & 0.00 & 1.90 & 0.08 & 11.00 & 34.00 & 1.00 & 3.51 & 0.56 & 9.40 \\ 
   \hline
\end{tabularx}
\caption{The fifth observation in wine quality dataset. D denotes density, $\text{Cl}^{-}$ stands for chlorides, "res." for residual and $\text{SO}_{4}^{2-}$ for sulphates.}\label{table:example_wine}
\end{table}



The SVM model used in this example is trained with the use of \pkg{e1071} package.
\begin{lstlisting}
library("e1071")
wine_svm_model <- svm(quality~., data = wine)
\end{lstlisting}

Different approaches for explaining single predictions are illustrated on the basis of the prediction for the fifth wine from this dataset, the one presented in table \ref{table:example_wine}. The actual quality of this wine is 5, while the quality predicted by the SVM model is $5.03$.

\begin{lstlisting}
nobs <- wine[5, ]
predict(wine_svm_model, nobs)
##        1 
## 5.032032
\end{lstlisting}

\subsection{The \pkg{live} package}

The \pkg{live} package approximates black box model (here SVM model) with a simpler white box model (here linear regression model) to explain the local structure of a black box model and in consequence to assess how features contribute to a single prediction.

To do this, we first need to generate artificial observations around the selected observation $x^{new}$ for local exploration.
We use \texttt{sample\_locally} function from \pkg{live} package.
\begin{lstlisting}
library("live")
library("mlr")
similar <- sample_locally(data = wine,
                          explained_instance = wine[5, ],
                          explained_var = "quality",
                          size = 500)
similar <- add_predictions(data = wine,
                           to_explain = similar,
                           black_box_model = wine_svm_model)
\end{lstlisting}

If multiple models are to be explained, there is no need to generate multiple \textit{artificial} datasets.
Predictions of each model on a single simulated dataset can be added with the use of \texttt{add\_predictions} function.
A different object should be created for each model, but the same result of a call to \texttt{sample\_locally} function should be used as the \texttt{to\_explain} argument.
Black box model can be passed as a model object or as a name of \pkg{mlr} learner.
While the object created by \texttt{sample\_locally} function stores the dataset and the name of the response variable, the object returned by \texttt{add\_predictions} function also stores the fitted black box model.
The result of applying \texttt{sample\_locally} functions does not contain the response but the result of \texttt{add\_predictions} contains a column with model predictions, which has the same name as response in the original dataset.

Once the artificial data points around $x^{new}$ are generated, we may fit the white box model to them. In this example we fit a linear regression model using \texttt{fit\_explanation} function.
\begin{lstlisting}
wine_expl <- fit_explanation(live_object = similar,
                             white_box = "regr.lm")
\end{lstlisting}
This function returns a native \texttt{mlr} object.
The model object (for example, \texttt{lm} object) can be extracted with the use of \texttt{getLearnerModel} function.

\begin{figure}
  \centering
  \includegraphics[width = 0.8\textwidth]{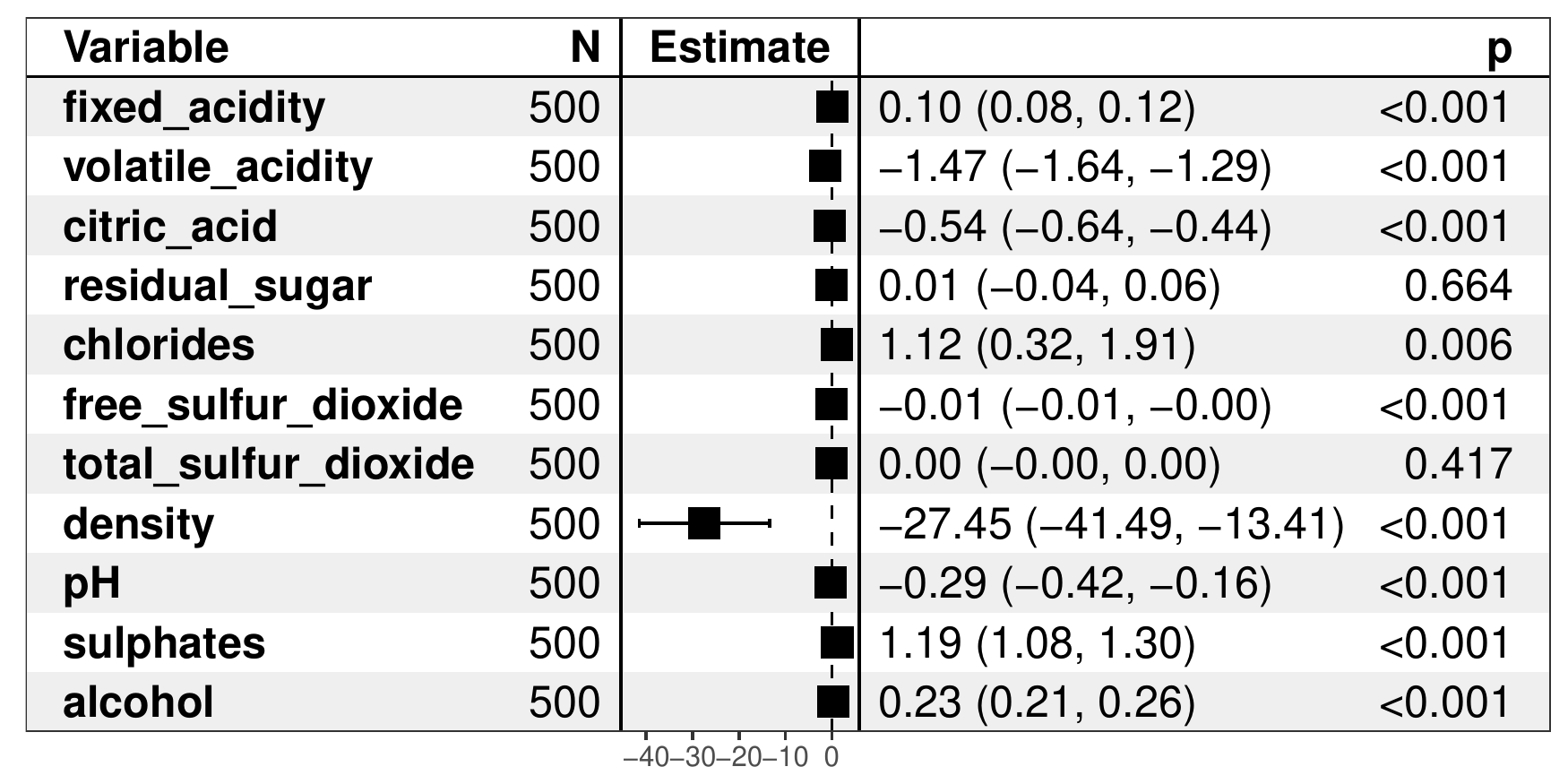}
  \caption{\label{fig:forestModel}\textit{Forest plot} for a linear white box model that approximates the black box model.}
  \includegraphics[width = 0.8\textwidth]{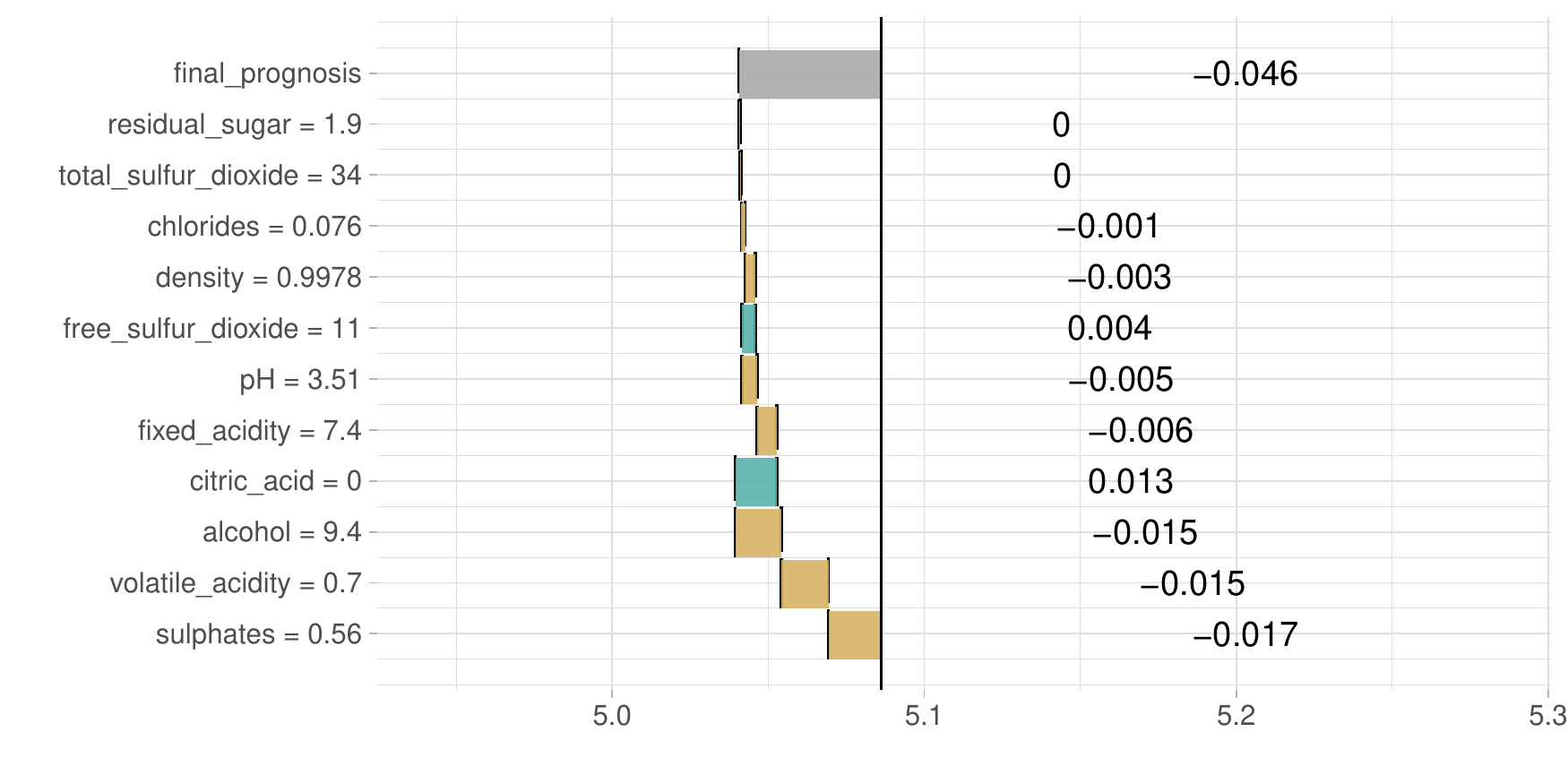}
  \caption{\label{fig:waterfallModel}\textit{Waterfall plot} for additive components of linear model that approximates the black box model around $x^{new}$.}
\end{figure}

The white box model \verb|wine_expl| approximates the black box model \verb|wine_svm_model| around $x^{new}$. Coefficients of this model can be presented graphically with the \texttt{plot\_explanation} function. See the corresponding Forest Plot in Figure \ref{fig:forestModel} and the corresponding Waterfall Plot in Figure \ref{fig:waterfallModel}.

\begin{lstlisting}
plot_explanation(model = wine_expl, regr_plot_type = "forestplot",
                 explained_instance = wine[5, ])
plot_explanation(model = wine_expl, regr_plot_type = "waterfallplot",
                 explained_instance = wine[5, ])
\end{lstlisting}

In case of datasets with larger number of variables, we could obtain sparse results by setting \texttt{selection = TRUE} in the \texttt{fit\_explanation} function. 
This option allows for performance of variable selection based on LASSO implemented in \pkg{glmnet} (\citep{lasso_pkg}, \citep{lasso_art}).
When using Generalized Linear Model as a white box model it is possible to set \texttt{family} argument to one of the distribution families available in \texttt{glm} and \texttt{glmnet} functions via \texttt{response\_family} argument to \texttt{fit\_explanation}.

\clearpage

\subsection{The \pkg{lime} package}

The LIME method is implemented in the R package \pkg{lime} \citep{lime_pkg}. It produces sparse explanations by default. 

In the first step a \texttt{lime} object is created for a specified dataset and a fitted black box model.
\begin{lstlisting}
library("lime")
wine_expl <- lime(nobs, wine_svm_model)
\end{lstlisting}
Then we use the \texttt{explain} function, which in case of regression takes the observation of interest, \texttt{lime} object and the number of top features to be used for explanation. 
In this case data is low dimensional, hence we can use all predictors.
Alternatively, we could set \texttt{feature\_select} to \texttt{none} to skip the selection part.

\begin{figure}[h!]
  \centering
  \includegraphics[width = 0.8\textwidth]{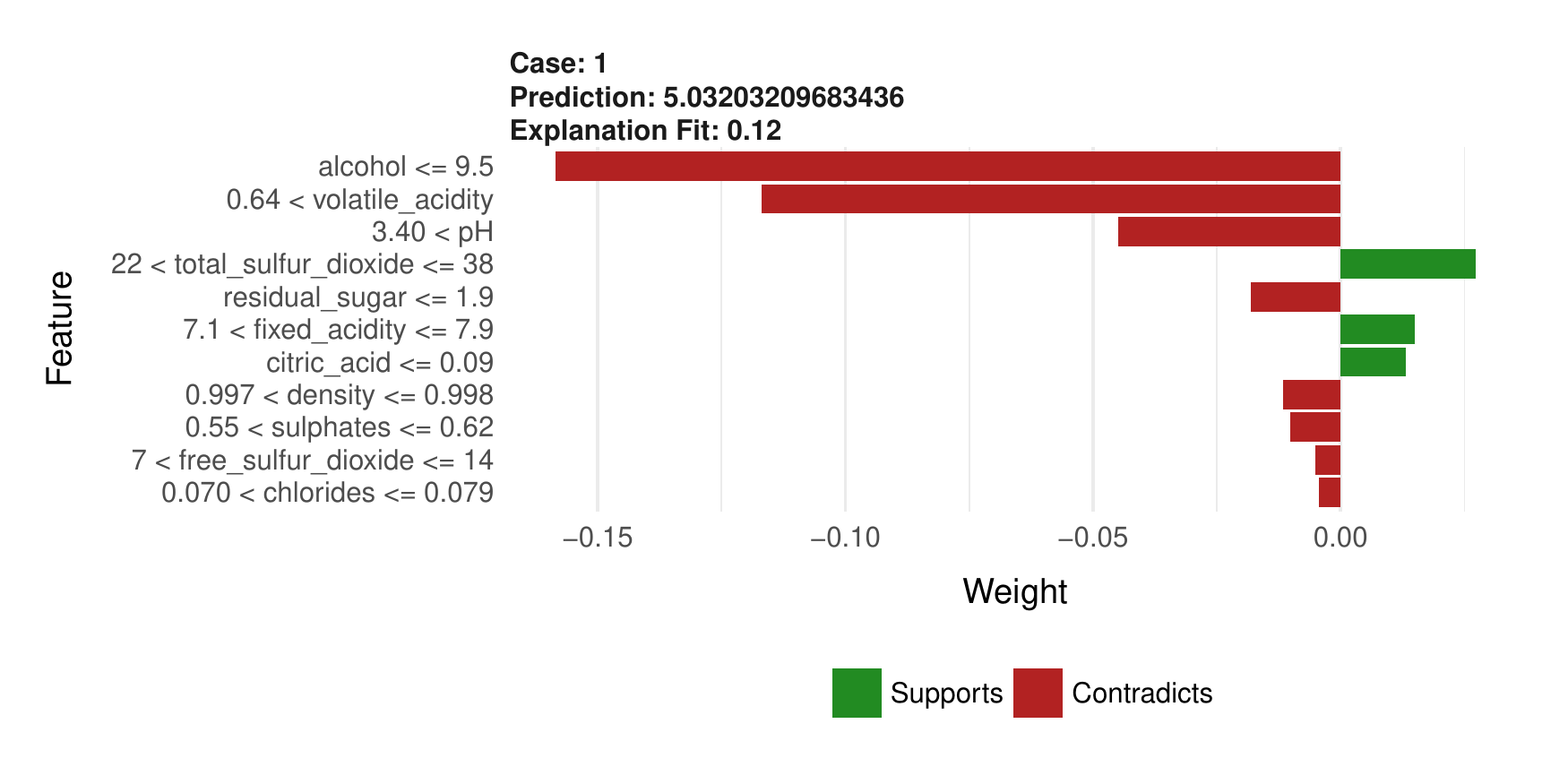}
  \caption{\label{fig:lime}Contributions of particular features to the prediction calculated with SVM model assessed with \pkg{lime} package.}
\end{figure}

\begin{lstlisting}
model_type.svm <- function(x, ...) "regression"
svm_explained <- explain(nobs, wine_expl, n_features = 11)
plot_explanation(svm_explained)
\end{lstlisting}
Results produced by the \texttt{plot\_explanation} function are presented in Figure \ref{fig:lime}.

\clearpage

\subsection{The \pkg{breakDown} package}

The \pkg{breakDown} package directly calculates variable attributions for a selected observation. It does not use any surrogate model.

The \texttt{broken()} function is used to calculate feature attributions. Generic functions \texttt{print()} and \texttt{plot()} show feature attributions as texts or waterfall plots.
The \texttt{baseline} argument specifies the origin of a waterfall plot. By default, it is 0. Use \texttt{baseline = "intercept"} to set the origin to average model prediction.

\begin{lstlisting}
library("breakDown")
explain_5 <- broken(wine_svm_model, new_observation = nobs, 
                    data = wine, 
                    baseline = "intercept",
                    direction = "up")
explain_5
##                             contribution
## baseline                           5.613
## + alcohol = 9.4                   -0.318
## + volatile_acidity = 0.7          -0.193
## + sulphates = 0.56                -0.068
## + pH = 3.51                       -0.083
## + residual_sugar = 1.9            -0.035
## + density = 0.9978                -0.031
## + chlorides = 0.076               -0.021
## + total_sulfur_dioxide = 34       -0.003
## + quality = 5                      0.000
## + free_sulfur_dioxide = 11         0.004
## + fixed_acidity = 7.4              0.024
## + citric_acid = 0                  0.144
## final_prognosis                    5.032

plot(explain_5)
\end{lstlisting}

\begin{figure}
\centering
\includegraphics[width = 0.8\textwidth]{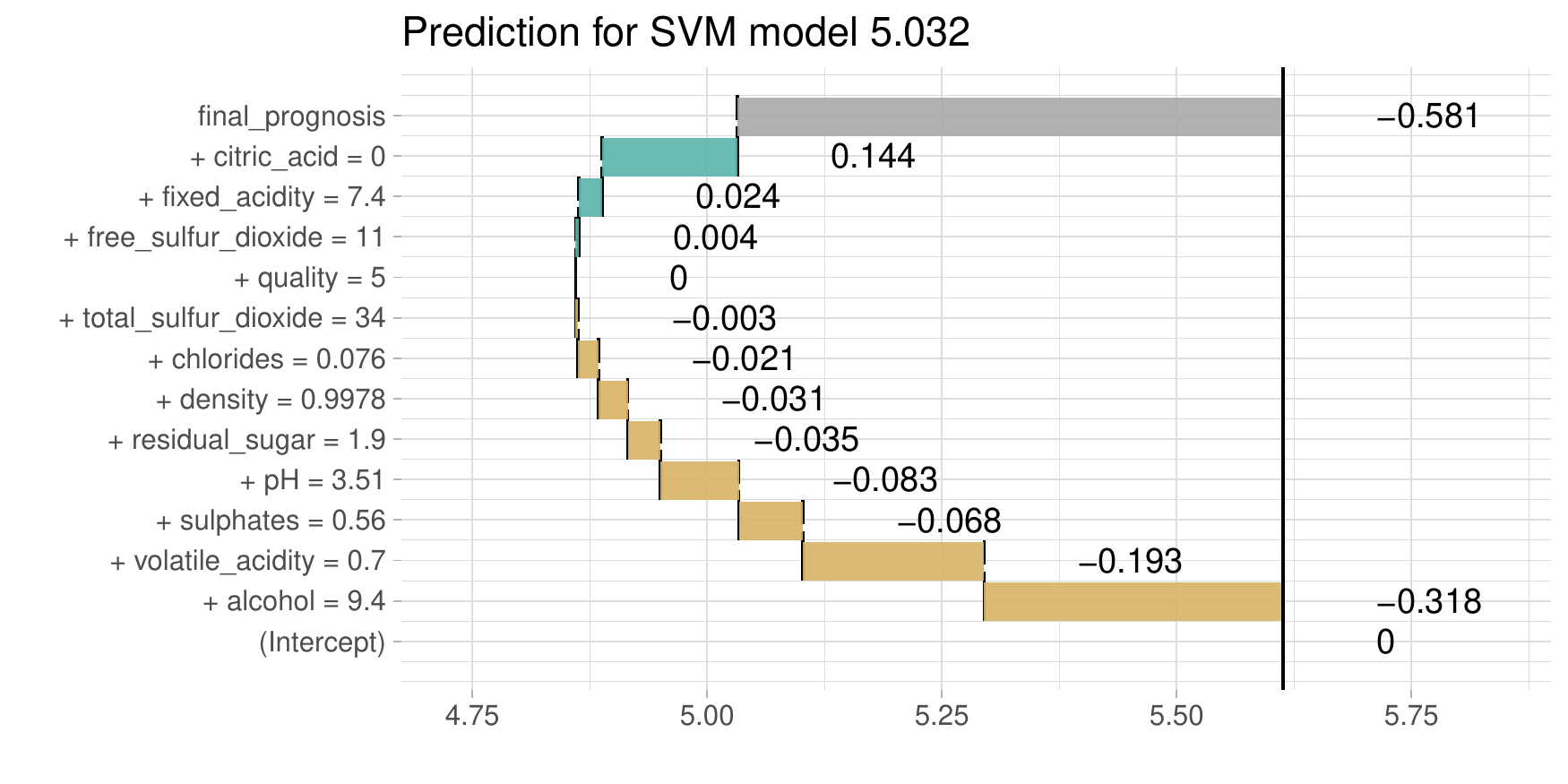}
\includegraphics[width = 0.8\textwidth]{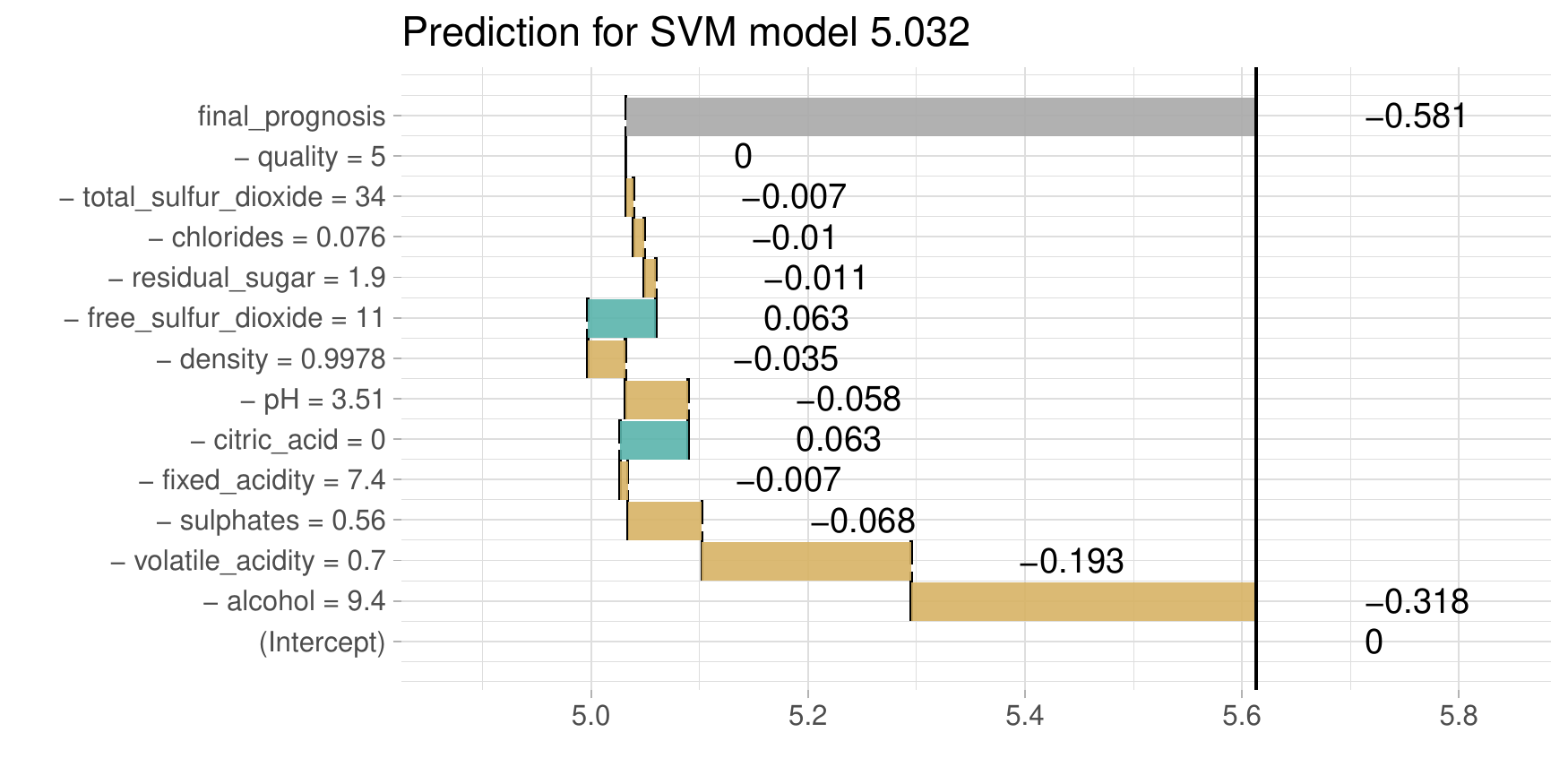}
\caption{\label{fig:breakDownSVN}\textbf{ag-break} feature attributions for SVM model calculated for the 5th wine. The upper plot presents feature attributions for the step-up strategy, while the bottom plot presents results for the step-down strategy. Attributions are very similar even if the ordering is different. Vertical black line shows the average prediction for the SVM model. The 5th wine gets final prediction of 5.032 which is below the average for this model by 0.581 point.}
\end{figure}

Figure \ref{fig:breakDownSVN} shows variable contributions for step-up and step-down strategy. Variable ordering is different but the contributions are consistent across both strategies.

Find more examples for classification and regression models created with \pkg{caret}, \pkg{mlr}, \pkg{randomForest} and other frameworks in package vignettes at \verb|https://pbiecek.github.io/breakDown/|.

\clearpage

\subsection{Shapley values (SHAP)}

Authors of the original article about Shapley values maintain a Python package which implements several methods of computing the Shapley values and provides visual diagnostic tools that help understand black box models such as plotting Shapley values for all instances in the dataset and exploring the influence of a single variable on predictions by plotting its value vs Shapley values and interaction Shapley values.

The only R package for Shapley values is at the development stage. It can be found at the address 
\url{https://github.com/redichh/ShapleyR}. 
So far, it allows user to compute Shapley values.
It uses the \texttt{mlr} interface to train models, what means that first, we need to create \texttt{mlr} task, and then pass it to \texttt{shapley} function along with the row number of an observation for which we explain the prediction and an \texttt{mlr} object with fitted model. 
\begin{lstlisting}
library(shapleyr)
tsk <- makeRegrTask("wine", wine, "quality")
shp <- shapley(5, model = train("regr.svm", tsk), task = tsk)
\end{lstlisting}
In table \ref{tab:comparisionSBD} we present Shapley values and compare them to contribution calculated with \textbf{ag-break} algorithm.

\begin{table}[ht]\label{shap}
\centering
\begin{tabular}{rlrr}
  \hline
 & Variable & Shapley values & \textbf{ag-break} contributions \\ 
  \hline 
  & Baseline prediction & 5.61 & 5.61 \\ 
1 & alcohol & -0.22 & -0.32 \\ 
  2 & sulphates & -0.19 & -0.07 \\ 
  3 & volatile\_acidity & -0.16 & -0.19 \\ 
  4 & citric\_acid & 0.13 & 0.14 \\ 
  5 & pH & -0.08 & -0.08 \\ 
  6 & fixed\_acidity & -0.04 & 0.02 \\ 
  7 & free\_sulfur\_dioxide & 0.03 & 0.00 \\ 
  8 & residual\_sugar & -0.02 & -0.03 \\ 
  9 & total\_sulfur\_dioxide & -0.01 & -0.00 \\ 
  10 & density & -0.01 & -0.03 \\ 
  11 & chlorides & 0.01 & -0.02 \\ 
  & Final prediction & 5.032 & 5.032 \\
   \hline
\end{tabular}
\caption{\label{tab:comparisionSBD}Comparison of feature attributions calculated with \pkg{ShapleyR} and \pkg{breakDown} packages.}
\end{table}

\clearpage

\section{Discussion}

In this paper we presented four approaches and four R packages that can be used for explanations of predictions from complex black box models. Two of them have already been introduced in literature, while \pkg{live} and \pkg{brakDown} were introduced in this article for the first time.

All four approaches are model agnostic in a sense that, the method does not depend on any particular structure of black box model. Yet there are also some differences between these approaches.

\begin{itemize}
\item Surrogate models vs. conditional expected responses. \pkg{live} and \pkg{lime} packages use surrogate models (the so-called white box models) that approximate local structure of the complex black box model. Coefficients of these surrogate models are used for explanations. Unlike them, \pkg{breakDown} and \pkg{shapleyr} construct feature attributions on the basis of conditional responses of a black box model.
\item \pkg{live} and \pkg{lime} packages differ in terms of the manner in which the surrounding of $x^{new}$ is defined. This task is highly non-trivial especially for mixed data with continuous and categorical features. \pkg{live} does not use \textit{interpretable input space} (and so does not fall under the \textit{additive feature attribution methods} category), but approximates the black box model directly in the data space, what can be considered a more effective use of data.
It comes with no theoretical guarantees that are provided for \textit{Shapley values}, but apart from being very intuitive, it offers several tools for visual inspection of the model.
\item \pkg{shapleyr} and \pkg{breakDown} take conditional expectation of the predictor function with respect to explanatory features. They differ in terms of the manner in which conditioning is applied to calculating feature attributions. \pkg{shapleyr} is based on results from game theory; in this package contribution of a single feature is averaged across all possible conditionings. \pkg{breakDown} uses a greedy approach in which only a single series of nested conditionings is considered. The greedy approach is easier to interpret and faster to compute. 
Moreover, exact methods of computing \textit{Shapley values} exist only for linear regression and tree ensemble models.
Approximate computations are also problematic, as they require the choice of number of samples of subsets of predictors which will be used.
These two methods produce nearly identical results for linear models (see table Table \ref{tab:comparisionSBD}), but for more complex models the estimated contributions can be very different, even pointing in opposite directions. An advantage of \textit{Shapley values} are proven theoretical properties, though they are restricted to explanation models that belong to the \textit{additive feature attribution methods} class.
\item When parameters (kernel and regularization term) are chosen as in \cite{shapley}, \pkg{lime} produces estimates of \textit{Shapley values}, while other choices of kernel and penalty term lead to inconsistent results.
The fact that the suggested penalty term is equal to 0 can be considered a huge limitation of LIME and SHAP, because in this setting they will not produce sparse explanations.
\item All presented methods decompose final prediction into additive components attributed to particular features. This approach will not work well for models with heavy components related to interactions between features.
\end{itemize}

Comparison of the methods presented in the previous section is far from being comprehensive. More research is needed to better understand differences between these approaches and new approaches are needed to overcome constraints listed above.
Nevertheless, the availability of the mentioned packages creates an opportunity for further studies on model exploration.

\section{Acknowledgements}

This work was financially supported by the \emph{NCN Opus grant 2016/21/B/ST6/02176}.

\nocite{*}
\bibliography{live}

\address{Mateusz Staniak \\
  Mathematical Institute\\
  University of Wroc{l}aw\\
  Poland \\
  \email{mateusz.staniak@math.uni.wroc.pl}}

\address{Przemysław Biecek \\
  Faculty of Mathematics and Information Science\\
	Warsaw University of Technology\\
    Faculty of Mathematics,  
    Informatics and Mechanics \\
    University of Warsaw\\
  Poland \\
ORCiD: 0000-0001-8423-1823\\
  \email{przemyslaw.biecek@gmail.com}}

\end{article}
\end{document}